\documentclass{article}
\pdfoutput=1
\usepackage{iclr2027_conference,times}
\iclrfinalcopy
\usepackage[T1]{fontenc}
\usepackage{amsmath,amssymb,booktabs,graphicx,microtype,url,multirow}
\usepackage[hidelinks]{hyperref}

\newcommand{\ContrastAllHi}{2.297}
\newcommand{\ContrastAllLo}{1.007}
\newcommand{\ContrastAllN}{8}
\newcommand{\ContrastAllSigN}{6}
\newcommand{\ContrastHi}{1.623}
\newcommand{\ContrastLo}{1.007}
\newcommand{\ContrastMedian}{1.261}
\newcommand{\ContrastNoKFourOneN}{6}
\newcommand{\ContrastSigN}{4}
\newcommand{\CurveHLNineM}{1.392}
\newcommand{\CurveHLOneTwoM}{1.257}
\newcommand{\CurveHLSixM}{1.098}
\newcommand{\CurveHLThreeM}{1.086}
\newcommand{\CurveSlope}{+0.0270}
\newcommand{\CurveSlopeP}{0.28}
\newcommand{\DepthNLFive}{300}

\newcommand{\DepthOrderP}{0.042}
\newcommand{\DepthRatioLFive}{0.384}
\newcommand{\DepthRatioLOneNine}{0.078}
\newcommand{\DepthRatioLOneTwo}{0.360}
\newcommand{\DepthRatioLTwoFour}{0.021}
\newcommand{\DepthSpan}{18}
\newcommand{\DispN}{3600}

\newcommand{\FireMax}{367}
\newcommand{\FireMedian}{103}
\newcommand{\FireMin}{31}

\newcommand{\GSvsRandHL}{2.269}
\newcommand{\GSvsRandHi}{2.655}
\newcommand{\GSvsRandLo}{1.933}
\newcommand{\GSvsRandN}{240}

\newcommand{\GThreeFixedN}{32}
\newcommand{\GThreeMinGatedRetained}{85}
\newcommand{\GThreeNewCons}{0.084}
\newcommand{\GThreeNewN}{38}
\newcommand{\GThreeNewPctAb}{32.3}
\newcommand{\GThreeOldCons}{0.155}
\newcommand{\GThreeOldPctAb}{10.6}
\newcommand{\GThreeSFifteenThirtySixFixedGain}{+0.176}
\newcommand{\GThreeSFifteenThirtySixGain}{+0.362}

\newcommand{\GThreeSFifteenThirtySixTopAgree}{9.4}
\newcommand{\GThreeSFifteenThirtySixUncErho}{0.291}

\newcommand{\GThreeSNinetySixFixedGain}{+0.046}
\newcommand{\GThreeSNinetySixGain}{+0.046}

\newcommand{\GThreeSNinetySixTopAgree}{13.6}
\newcommand{\GThreeSNinetySixUncErho}{0.320}

\newcommand{\GThreeSThreeEightyFourFixedGain}{+0.082}
\newcommand{\GThreeSThreeEightyFourGain}{+0.138}
\newcommand{\GThreeSThreeEightyFourGainHi}{+0.290}
\newcommand{\GThreeSThreeEightyFourGainLo}{+0.024}

\newcommand{\GThreeSThreeEightyFourJaccard}{0.121}
\newcommand{\GThreeSThreeEightyFourPairedN}{53}
\newcommand{\GThreeSThreeEightyFourPerArmErho}{0.634}
\newcommand{\GThreeSThreeEightyFourPerArmPctAb}{11.9}
\newcommand{\GThreeSThreeEightyFourSharedErho}{0.767}
\newcommand{\GThreeSThreeEightyFourSharedPctAb}{2.4}
\newcommand{\GThreeSThreeEightyFourTopAgree}{11.3}
\newcommand{\GThreeSThreeEightyFourUncErho}{0.350}

\newcommand{\GThreeSThreeEightyFourUncPctAb}{24.8}

\newcommand{\GThreeUnionGain}{+0.037}
\newcommand{\GThreeUnionGainHi}{+0.171}
\newcommand{\GThreeUnionGainLo}{-0.046}
\newcommand{\GThreeUnionN}{36}
\newcommand{\GThreeUnionPctAb}{6.3}
\newcommand{\GThreeUnionRetained}{66}

\newcommand{\GTwoDispNullMedian}{137}
\newcommand{\GTwoDispNullSame}{3.5}

\newcommand{\GTwoDispObsMedian}{118}
\newcommand{\GTwoDispObsSame}{13.9}
\newcommand{\GTwoFixedN}{62}
\newcommand{\GTwoMinGatedRetained}{131}
\newcommand{\GTwoNewCons}{0.105}

\newcommand{\GTwoNewPctAb}{18.8}
\newcommand{\GTwoOldCons}{0.114}
\newcommand{\GTwoOldPctAb}{8.7}
\newcommand{\GTwoSFifteenThirtySixFixedGain}{+0.110}
\newcommand{\GTwoSFifteenThirtySixGain}{+0.186}

\newcommand{\GTwoSFifteenThirtySixTopAgree}{10.0}
\newcommand{\GTwoSFifteenThirtySixUncErho}{0.469}

\newcommand{\GTwoSNinetySixFixedGain}{+0.082}
\newcommand{\GTwoSNinetySixGain}{+0.082}

\newcommand{\GTwoSNinetySixJaccard}{0.158}

\newcommand{\GTwoSNinetySixTopAgree}{18.1}
\newcommand{\GTwoSNinetySixUncErho}{0.433}

\newcommand{\GTwoSThreeEightyFourFixedGain}{+0.110}
\newcommand{\GTwoSThreeEightyFourGain}{+0.130}
\newcommand{\GTwoSThreeEightyFourGainHi}{+0.208}
\newcommand{\GTwoSThreeEightyFourGainLo}{+0.074}

\newcommand{\GTwoSThreeEightyFourJaccard}{0.141}
\newcommand{\GTwoSThreeEightyFourPairedN}{93}
\newcommand{\GTwoSThreeEightyFourPerArmErho}{0.738}
\newcommand{\GTwoSThreeEightyFourPerArmPctAb}{7.6}
\newcommand{\GTwoSThreeEightyFourSharedErho}{0.869}
\newcommand{\GTwoSThreeEightyFourSharedPctAb}{0.0}
\newcommand{\GTwoSThreeEightyFourTopAgree}{13.9}
\newcommand{\GTwoSThreeEightyFourUncErho}{0.492}

\newcommand{\GTwoSThreeEightyFourUncPctAb}{18.4}

\newcommand{\GTwoUnionGain}{+0.085}
\newcommand{\GTwoUnionGainHi}{+0.151}
\newcommand{\GTwoUnionGainLo}{+0.037}
\newcommand{\GTwoUnionN}{64}
\newcommand{\GTwoUnionPctAb}{0.0}
\newcommand{\GTwoUnionRetained}{118}

\newcommand{\MatchedTauEightyHL}{1.114}
\newcommand{\MatchedTauEightyHi}{1.270}
\newcommand{\MatchedTauEightyLo}{0.982}
\newcommand{\MatchedTauNinetyHL}{1.356}
\newcommand{\MatchedTauNinetyHi}{1.581}
\newcommand{\MatchedTauNinetyLo}{1.241}
\newcommand{\MineN}{240}
\newcommand{\MineVsFrozenHL}{1.192}
\newcommand{\MineVsFrozenHi}{1.386}
\newcommand{\MineVsFrozenLo}{1.023}
\newcommand{\MineVsFrozenP}{0.0213}
\newcommand{\MineVsUntrHL}{2.104}
\newcommand{\MineVsUntrHi}{2.415}
\newcommand{\MineVsUntrLo}{1.836}

\newcommand{\NDicts}{9}
\newcommand{\NormFlipHL}{1.687}
\newcommand{\NormFlipHi}{2.498}
\newcommand{\NormFlipLo}{1.171}

\newcommand{\NormFlipP}{0.011}
\newcommand{\PNormFreq}{14.76}
\newcommand{\PNormRare}{3.60}

\newcommand{\PosResidRTwo}{0.005}

\newcommand{\RateDtopNegPct}{46}
\newcommand{\RateFlipMax}{0.110}
\newcommand{\RateFlipMin}{0.017}
\newcommand{\RateN}{1500}
\newcommand{\RateRatio}{3.3}
\newcommand{\RateRhoAdtop}{0.232}
\newcommand{\RateRhoDtopSigned}{0.022}
\newcommand{\RateRhoFlip}{0.100}
\newcommand{\RateRhoKL}{0.328}
\newcommand{\RateRhoKLRaw}{0.295}
\newcommand{\RawFlipHL}{0.683}
\newcommand{\RawFlipHi}{0.961}
\newcommand{\RawFlipLo}{0.459}
\newcommand{\RawFlipNR}{35}
\newcommand{\RawFlipNT}{40}
\newcommand{\RawFlipP}{0.039}
\newcommand{\RelSparsityBandAgree}{11.2}

\newcommand{\RelSparsityNearAgree}{42.7}
\newcommand{\RelSparsityNearJaccard}{0.404}
\newcommand{\RelSparsityNearMedCos}{0.90}

\newcommand{\RelSparsityTopAgree}{60.2}

\newcommand{\RelSparsityWeighted}{42}
\newcommand{\RelWidthBandAgree}{10.5}

\newcommand{\RelWidthNearAgree}{38.8}
\newcommand{\RelWidthNearJaccard}{0.361}
\newcommand{\RelWidthNearMedCos}{0.87}

\newcommand{\RelWidthTopAgree}{52.3}

\newcommand{\RelWidthWeighted}{35}
\newcommand{\RhoCIHalf}{0.10}
\newcommand{\RhoContrastP}{2.1e-06}
\newcommand{\RhoContrastZ}{4.74}
\newcommand{\RhoDeep}{0.088}
\newcommand{\RhoLFive}{0.358}
\newcommand{\RhoLOneNine}{0.031}
\newcommand{\RhoLOneTwo}{0.339}
\newcommand{\RhoLTwoFour}{0.145}
\newcommand{\RhoShallow}{0.349}
\newcommand{\RhoSlopePerLayer}{-0.0163}
\newcommand{\SDRatioAll}{42.1}
\newcommand{\SDRatioDropOne}{7.2}
\newcommand{\ScopeAgreeMax}{48}
\newcommand{\ScopeAgreeMin}{30}
\newcommand{\ScopeCosMax}{0.91}
\newcommand{\ScopeCosMin}{0.78}
\newcommand{\ScopeCosPairMedian}{0.854}
\newcommand{\ScopeNDicts}{6}
\newcommand{\ScopePairs}{15}
\newcommand{\ScopeShareLow}{3.9}
\newcommand{\ScopeShareTop}{34.7}

\newcommand{\SixArmErhoTwo}{0.415}
\newcommand{\SixArmI}{200}
\newcommand{\SixArmJ}{6}
\newcommand{\SixArmN}{6}

\newcommand{\SixArmPctArm}{3.2}
\newcommand{\SixArmPctInter}{12.5}
\newcommand{\SixArmPctLatent}{16.9}
\newcommand{\SixArmPctResid}{67.4}

\newcommand{\SixArmSharedN}{240}
\newcommand{\TailMassMax}{73}
\newcommand{\TailMassMin}{36}
\newcommand{\TailMaxMedMax}{464}
\newcommand{\TailMaxMedMin}{33}

\newcommand{\TwoArmErhoTwo}{0.455}
\newcommand{\TwoArmPctArm}{0.4}
\newcommand{\TwoArmPctInter}{9.7}

\newcommand{\codeavailability}{Code and data: \url{https://github.com/vcnoel/sae-artifact}.}

\title{Where You Measure Decides\\
       What You Measure: Position Selection in Ablation-Based SAE Evaluation}
\author{
  Valentin No\"el \\
  Devoteam \\
  \texttt{valentin.noel@devoteam.com}
}

\begin{document}
\maketitle
\lhead{}\chead{}\rhead{}

\begin{abstract}
Sparse autoencoders are meant to name the things a language model computes, and
the usual way to check that a latent matters is to switch it off and see what
changes. But a latent fires at many tokens, and the effect has to be measured at
one of them. The convention is to measure where the latent fires hardest. That
choice is almost never reported, and it is not made by the experimenter: it is
made by the dictionary under evaluation. Change the dictionary and the
measurement moves to a different token.

We show this is not a detail. Take two sparse autoencoders released by Google
for the same model and match their latents by decoder similarity: even among the
pairs the two dictionaries encode almost identically, they pick different tokens
for a large share of them. Two dictionaries compared under the usual protocol are
therefore very often compared at different places. To separate the
convention from the dictionaries we train six autoencoders from one
initialisation, differing only in fitting choices, so that a latent means the
same thing in each. Most of the variance such a comparison reads as
\emph{these dictionaries disagree about this latent} turns out to be the
position instead: it falls from $\GTwoSThreeEightyFourPerArmPctAb\%$ and
$\GThreeSThreeEightyFourPerArmPctAb\%$ of variance to near zero once every
dictionary is measured at the same token. More evaluation data does not rescue
it. Across a sixteenfold range of corpus sizes the dictionaries agree
less about where to measure, not more, so the problem grows with scale.

The correction is one line of evaluation code. We give the protocol an
ablation-based causal number must report to be comparable across papers, and an
audit of five published papers against it. In short: a causal number reported
without its position
describes the token it was taken at as much as the latent it was taken from.

\codeavailability
\end{abstract}

\section{Introduction}
Sparse autoencoders (SAEs) are the current tool of choice for turning a
language model's activations into a list of named parts
\citep{bricken2023monosemanticity, cunningham2023sparse, templeton2024scaling}.
Once you have such a list, the obvious next question is which parts matter, and
the obvious way to answer it is intervention: switch a latent off, run the model
again, and read how much the output changed
\citep{marks2024sparse, gao2024scaling, cho2026singletoken}. The resulting
number is written down as a property of the latent, and read as one.

There is a step in the middle that nobody writes down. A latent does not fire
once; it fires at many tokens across many sequences, and the intervention has to
happen at one of them. Which one? The near-universal answer is the token where
the latent fires hardest: a sensible instinct, inherited from single-neuron
neuroscience, that the way to characterise a unit is to find what drives it
most. The difficulty is that the token where a latent fires hardest is not a
fact about the model. It is a fact about the dictionary, computed from the
dictionary's own activations. Fit a second autoencoder on the same data and it
will place the same latent's maximum somewhere else.

That makes the measurement location a downstream consequence of the very thing a
comparison is trying to evaluate. When two papers report causal numbers for
comparable latents, or when one paper compares two dictionaries, the numbers are
not measured at the same place, and nothing in either paper says so, because
the convention is invisible enough that it is rarely stated at all.

This paper measures how much that costs. We do it first on dictionaries we had
no hand in: two Gemma Scope autoencoders released for the same base model, where
we can match latents across dictionaries and ask how often they would be
measured at the same token. Then, to separate the convention from the
dictionaries themselves, we train six autoencoders from one shared
initialisation that differ only in fitting choices, so that a latent denotes the
same thing in all six and any disagreement between them is attributable rather
than confounded. Measuring each dictionary where it prefers, and then measuring
all of them at one common token, isolates what the convention contributes.

The answer is that it contributes most of it. What a conventional comparison
reads as these dictionaries disagree about this latent is largely the
dictionaries being read at different tokens, and holding the token fixed removes
almost all of it. This does not improve with a bigger evaluation corpus; it gets
worse, because more text means more places for two dictionaries to disagree
about. The correction is a single line of evaluation code.

Two smaller reporting choices behave the same way. Whether special-token
positions are dropped, and whether the readout is normalised by the size of the
intervention, can each flip the sign of a real comparison on identical
data (\S\ref{sec:protocol}), and neither is usually written down. We surveyed five
published papers that zero-ablate a single latent and read a magnitude, and
recorded what each states about these conventions. The survey asks what the
papers report; it cannot ask what their authors did, and we would expect
some of these controls to have been applied and simply not described. That gap
is the point rather than a complaint: our own earlier draft used the
top-activating convention without reporting it, for the same reason everyone
does: it did not look like a choice.

\paragraph{What it would take to falsify this.} Take any two sparse
autoencoders on the same base model. Record where each would measure a given
latent under the top-activating convention: if they agree, our account is
wrong. Then decompose the latent$\times$arm variance with each dictionary at
its own positions and again at a shared set: if that component does not fall,
the repair does not work. Both run on one GPU in an afternoon, need no training
beyond dictionaries a reader already has, and are a single flag in our
implementation. We report them on two base models, six arms, and three
evaluation corpora, and every number below carries the corpus it came from.

\textbf{Contributions:}

\begin{itemize}
\item The measurement position is selected by the dictionary under evaluation,
so two dictionaries are compared at different tokens. Released Gemma Scope
dictionaries pick different tokens for a large share of the latents they encode
almost identically, and in a controlled six-arm design, holding position
fixed collapses the latent$\times$arm variance component that a comparison would
otherwise attribute to the dictionaries (\S\ref{sec:property},
Appendix~\ref{app:released}).
\item Two further reporting choices, special-token handling and normalisation
by intervention magnitude, can flip the sign of a real comparison on identical
data (\S\ref{sec:protocol}, Appendix~\ref{app:flip}).
\item A protocol stating what an ablation-based causal number must report to be
comparable across papers (\S\ref{sec:protocol}), and an audit of five published
papers against it (Table~\ref{tab:audit}).
\end{itemize}

\section{Related work}
\label{sec:relwork}
\textbf{Dictionaries.} Sparse autoencoders decompose a residual stream into an
overcomplete set of sparsely-active directions
\citep{bricken2023monosemanticity, cunningham2023sparse}, motivated by
superposition: more features than dimensions, stored at non-orthogonal
directions \citep{elhage2022toy}, with the linear-representation view refined
since \citep{park2023linear, engels2024not}. The line has
since scaled to production models \citep{templeton2024scaling},
to TopK and JumpReLU objectives \citep{gao2024scaling,
rajamanoharan2024jumprelu}, to gated and batched variants
\citep{rajamanoharan2024improving, bussmann2024batchtopk}, to objectives that
select for downstream effect rather than reconstruction
\citep{braun2024identifying}, to other sites and to transcoders
\citep{kissane2024interpreting, dunefsky2024transcoders}, and to public
dictionary suites \citep{lieberum2024gemmascope}. We take these methods as
given: this paper measures a convention applied to such dictionaries, not the
dictionaries themselves.

\textbf{Evaluating dictionaries.} SAEBench \citep{karvonen2025saebench}
consolidated the metrics the field reports, and a run of results has since
questioned what they discriminate. \citet{paulo2025sparse} show that seed alone
substantially changes which features are learned, TopK more than ReLU
(\S\ref{sec:setup}); \citet{heap2025automated} find automated
interpretability scores do not separate trained transformers from random ones;
\citet{korznikov2026sanity} show frozen-random baselines match fully-trained
SAEs across reconstruction, interpretability, sparse probing, and causal
editing ($0.73$ vs $0.72$); and \citet{chanin2026reliable} finds SAEBench's TPP
and SCR fail three independent reliability lenses. Two of these sit outside our
audit table on scope rather than merit: Korznikov's causal-editing metric is
RAVEL's interchange match-rate, and TPP/SCR ablate a small set of
latents per concept, not one. Their mechanisms, reseed noise, low
discriminability across training trajectories, are complementary to the
directional confound in a fixed measurement that we document; together the
results say this literature currently cannot distinguish a trained dictionary
from a matched baseline with the tools on offer. Evaluations that supply
external ground truth are the partial exception
\citep{karvonen2024measuring, chaudhary2024evaluating,
kantamneni2025are}, and a second line reports that the units themselves are
unstable under width and sparsity, splitting and absorbing one another
\citep{leask2025sparse, chanin2024a}: our position confound is orthogonal to
both, since it survives holding the latent identity fixed by construction
(\S\ref{sec:position}).

\textbf{Ablation as a causal readout.} Zero-ablating a latent and reading a
magnitude is the standard single-latent causal probe \citep{marks2024sparse,
gao2024scaling, cho2026singletoken}, and is distinct from interchange
interventions scored by match-rate accuracy \citep{huang2024ravel,
makelov2024principled}, which our results do not address (\S\ref{sec:limitations}).
It inherits its logic from activation patching over model components
\citep{meng2022locating, wang2022interpretability}, which has been automated
\citep{conmy2023towards} and approximated by gradients for scale
\citep{syed2023attribution, kramar2024atp}. That literature has already found
that the choice of patching metric and corrupted baseline changes the
conclusion \citep{zhang2023towards}; the facet we isolate, which position the
intervention is applied at, is the same kind of degree of freedom one level
down, and it is chosen by the dictionary rather than the experimenter.
Downstream uses inherit whatever the measurement selects, since latents chosen
this way are what steering and unlearning interventions act on
\citep{turner2023steering, chalnev2024improving, farrell2024applying}.
\citet{cho2026singletoken} is the closest published methodology to ours. Their
headline depth correlation ($\rho=0.81$, their Table 21) survives a sensitivity
bound on within-layer latent variance with room to spare and we did not find
grounds to contest it; separately, our implementation of their single-token
detector at their exact specification (layer 12, Gemma-2-2B) yields
$0.41$--$0.44\times$ their reported prevalence at four times their apparent
corpus size: an unexplained reproducibility discrepancy we report about the
detector, not their causal claim.

\textbf{Measurement reliability.} The question this paper asks, how much of
an observed difference is the object and how much is the measurement procedure, is generalizability theory's \citep{cronbach1972dependability,
brennan2001generalizability, shavelson1991generalizability}, and the specific
failure we document is the one \citet{clark1973language} named: treating a
sampled facet as fixed. Psycholinguistics and social psychology absorbed that
correction by moving to crossed random-effects designs
\citep{baayen2008mixed, judd2012treating}. Machine learning has an equivalent
literature on variance in benchmark comparisons
\citep{bouthillier2021accounting, dodge2019show, henderson2017deep,
card2020with}, but it treats seeds and
splits as the sampled facets. We add one it has not: the measurement
position, which is unusual in being selected by the system under evaluation.
The closest precedent is \citet{bolukbasi2021an}, who show that reading a unit
off its maximally-activating inputs can manufacture an interpretation that does
not survive a change of context. Their illusion is about what the selected
inputs license one to \emph{say}; ours is about what the selected position does
to a \emph{number}.

\section{Setup}
\label{sec:setup}
We measure the causal effect of a single SAE latent by ablating its contribution
to the residual stream and reading the change in the model's next-token
distribution. For latent $f$ with decoder direction $d_f$ and activation $a_f$ at
position $t$, we replace $h_t \leftarrow h_t - a_f d_f$ at the SAE's hook point
and record
\begin{equation}
\mathrm{KL}_t = D_{\mathrm{KL}}\!\left(p(\cdot \mid h) \,\|\, p(\cdot \mid h')\right),
\label{eq:kl}
\end{equation}
evaluated at position $t$ and, for the propagation analysis, at $t{+}1$ and
$t{+}3$. Experiments use \texttt{gemma-2-2b} and residual-stream Gemma Scope
dictionaries, a matched trained/soft-frozen pair, and six SAE arms sharing a
single seed (\S\ref{sec:property}).

\paragraph{Why the arms share an initialisation.} The crossed design requires
that latent $i$ denote the same thing in every arm, and that is not something
SAE training provides on its own. \citet{paulo2025sparse} find that
autoencoders trained on identical data and differing only in random seed share
roughly $30\%$ of their features, and that TopK architectures, ours, are
more seed-dependent than ReLU ones. Seeding all six arms identically is
therefore not a convenience but the condition under which a latent-wise
comparison means anything at all. It also fixes the direction of our scope:
arms sharing an initialisation are the most similar dictionaries a practitioner
would ever compare, so the instability measured here is a lower bound on what
a comparison across seeds would show.

The six-arm design is run on two base
models, \texttt{gemma-2-2b} and \texttt{gemma-3-1b}, at the same relative depth
(layer $12$ of $26$ in both) and identical dictionary shape, and at two
evaluation corpora ($96$ and $384$ sequences). Every number below states which
model and which corpus produced it.

\paragraph{Estimator.}
Group comparisons use the Hodges--Lehmann estimator, the median of pairwise
ratios, reported with a bootstrap interval. HL is the point estimate belonging to
the Mann--Whitney test; pairing a Mann--Whitney $p$-value with a
ratio-of-medians interval, as we did initially, produces an apparent
contradiction that is an artifact of mixing frameworks.

\paragraph{Reliability.} We report the crossed latent$\times$arm design with the
statistic generalizability theory defines for it
\citep{cronbach1972dependability, brennan2001generalizability}. Writing
$v_a$ for the variance between latents, $v_{ab}$ for the latent$\times$arm
interaction, and $v_e$ for residual variance, the generalizability coefficient
over $n$ arms is
\begin{equation}
E\rho^{2} \;=\; \frac{v_a}{v_a + v_{ab} + v_e/n},
\label{eq:erho2}
\end{equation}
the expected squared correlation between a measurement taken on one set of arms
and one taken on another. It answers exactly the question a reader of a
published causal number needs answered, would this ranking of latents
survive a different dictionary?, and the interaction term $v_{ab}$ sits in
its denominator, which is why any convention that inflates $v_{ab}$ depresses
$E\rho^{2}$ arithmetically rather than as a separate finding. Components are
estimated by method of moments and clipped at zero, which biases $v_{ab}$
upward and therefore $E\rho^{2}$ downward in the controlled arm; we
report absolute components alongside every coefficient so the clipping is
visible.

\paragraph{Convention.}
Ratios of the form $\mathrm{KL}_{t+k}/\mathrm{KL}_t$ are computed
feature-level: a per-latent median first, then the ratio of medians across
latents. The observation-level alternative differs by a non-constant factor
$0.97$--$1.35$ and is not used.

\section{The measurement position is chosen by the dictionary}
\label{sec:property}
The sign flip of Appendix~\ref{app:flip} shows that how a single fixed dataset
is read can flip a conclusion. This section asks a different question: does the
same latent, measured on dictionaries that differ only in a fitting choice,
keep its rank at all? The answer turns out to depend almost entirely on one
convention that no audited paper reports, \emph{where} the latent is
measured, and the rest of this section separates that convention from the
dictionaries it is confounded with.

\begin{figure}[t]\centering
\includegraphics[width=\textwidth]{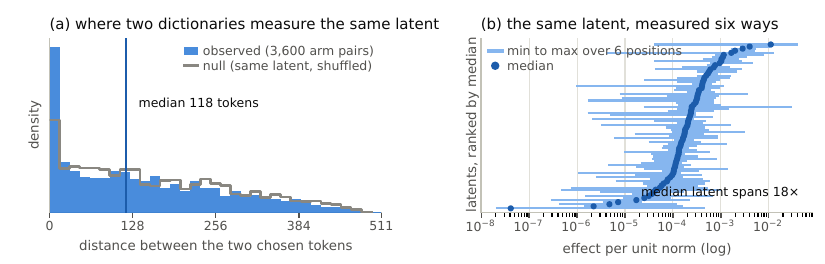}
\caption{(a) Distance between the tokens two dictionaries pick for the same
latent, over all $15$ arm pairs ($\DispN$ comparisons, Gemma-2-2B, $384$
sequences). The null draws each arm's choice from the positions that latent
actually fires at, since an arm can only pick where the latent fires. Agreement
is real but small: $\GTwoDispObsSame\%$ land on the same token against
$\GTwoDispNullSame\%$ under the null, roughly four times chance. Beyond that the
choices are near-independent (median separation $\GTwoDispObsMedian$ tokens
against $\GTwoDispNullMedian$). (b) Each line spans a latent's smallest and
largest effect across its six measured positions, dot at the median, latents
ranked by that median.}
\label{fig:mechanism}
\end{figure}

\paragraph{The convention, and why it is not innocuous.} A latent is
conventionally measured at the token where it fires hardest. That token is a
function of the dictionary: a different fit assigns different activations, so it
selects a different position. The measurement location is therefore not a
constant of the experiment but a downstream consequence of the thing being
compared. Across our six arms, the position sets selected for the same latent
overlap at Jaccard $\GTwoSThreeEightyFourJaccard$ (Gemma-2-2B) and
$\GThreeSThreeEightyFourJaccard$ (Gemma-3-1B), and only
$\GTwoSThreeEightyFourTopAgree\%$ and $\GThreeSThreeEightyFourTopAgree\%$ of
latents have the same single top-activating position in
any two arms. These are not stable constants of the method: at a quarter of this
evaluation corpus the same two Gemma-2-2B quantities read
$\GTwoSNinetySixJaccard$ and $\GTwoSNinetySixTopAgree\%$, and they should be
expected to
fall further as the corpus grows, since more sequences offer more candidate
positions for the arms to disagree about. The right statement is that
disagreement grows with corpus size, which makes the uncontrolled
convention worse at scale rather than better: the opposite of the usual
expectation that more evaluation data stabilises a measurement. Comparing a
latent's
measured effect across dictionaries under this convention compares it at
different tokens, and attributes the difference to the dictionary.

\paragraph{The dictionaries agree more than chance, and far less than a
comparison assumes.} The convention's name suggests a handful of clear
candidates, and it is worth seeing how far from that the data is: on this corpus
a released dictionary's latents are live at a median of $\FireMedian$ positions
each, ranging from $\FireMin$ to $\FireMax$ across the six dictionaries. ``The
token where it fires hardest'' is therefore a selection of one ticket from
dozens to hundreds, and which ticket was drawn is what no paper reports.

A reader who thinks about that denominator should reach the opposite reading
first: with so many live positions, is agreeing even one time in ten not
remarkable? It is above chance, and
the right comparison is the shuffle null of Figure~\ref{fig:mechanism}: drawing
each arm's choice at random from the positions that latent actually fires at
gives $\GTwoDispNullSame\%$ against the observed $\GTwoDispObsSame\%$, so the
arms agree about four times as often as chance. They are not picking arbitrary
tokens. That is what makes the residual gap a problem rather than a curiosity:
a published causal number is not offered as agreeing with another paper's
number a few times more often than chance, it is offered as measuring the same
thing, and the arms disagree about where to measure far too often for that.

\paragraph{What this does to the design.} Because each arm selects its own
positions, position is nested within (latent, arm) rather than crossed with
latent. The latent$\times$arm interaction then mixes two things that cannot be
separated after the fact: the latent behaving differently under a different fit,
and the arms having chosen different tokens. Only $49$ of $5{,}436$
(latent, position) cells were common to all six arms in our first measurement,
so the confound could not be resolved by subsetting: it required
re-measuring. We therefore report two designs throughout: \emph{per-arm}, in
which each arm uses its own top-activating positions, which is published
practice; and \emph{shared}, in which one arm-symmetric position set is chosen
per latent (candidates ranked by the minimum gated activation across arms, so
the latent demonstrably fires in every arm at every selected position, and no arm
serves as the reference) and every arm is measured there. The contrast between
them is the effect of the convention.

\paragraph{An earlier version of this paper reported the per-arm numbers
alone.} Its conclusion, that causal importance is not a property of the
latent, was measured under a protocol in which the dictionary chose the
measurement location, and it does not survive controlling that. The retraction
and its evidence are in the supplementary record; what follows is the controlled
result.

\paragraph{Design.} Six TopK SAEs share initialisation seed $0$, so latent $i$
denotes the same initial direction in every arm; each is then trained with one
different, individually defensible choice: decoder free, soft-frozen at
$\tau=0.80$, soft-frozen at $\tau=0.90$, learning rate $10\times$ lower, sparsity
$k=41$ instead of $82$, or a reshuffled corpus order. Every arm sees the same
$12$M tokens. Live latents are intersected across all six arms
($\SixArmSharedN$ shared), one uniform sample is drawn from the intersection,
and every arm is measured on exactly that sample at up to six positions per
latent -- giving a crossed latent$\times$arm design with real replication on
both factors, the direct six-level analogue of a model$\times$wrapper design.

\begin{table}[t]\centering
\begin{tabular}{llccc}
\toprule
model & positions & latents & latent$\times$arm & $E\rho^2$ \\
\midrule
Gemma-2-2B & per-arm & \GTwoSThreeEightyFourPairedN & $\GTwoSThreeEightyFourPerArmPctAb\%$ & $\GTwoSThreeEightyFourPerArmErho$ \\
Gemma-2-2B & shared  & \GTwoSThreeEightyFourPairedN & $\GTwoSThreeEightyFourSharedPctAb\%$  & $\GTwoSThreeEightyFourSharedErho$ \\
\addlinespace
Gemma-3-1B & per-arm & \GThreeSThreeEightyFourPairedN  & $\GThreeSThreeEightyFourPerArmPctAb\%$ & $\GThreeSThreeEightyFourPerArmErho$ \\
Gemma-3-1B & shared  & \GThreeSThreeEightyFourPairedN  & $\GThreeSThreeEightyFourSharedPctAb\%$  & $\GThreeSThreeEightyFourSharedErho$ \\
\bottomrule
\end{tabular}
\caption{The measurement position, not the dictionary, carries the
latent$\times$arm variance. Within each model the two rows use identical latents
at identical replication and differ only in where the measurement was taken.
Evaluation corpus $384$ sequences. The paired gain in $E\rho^2$ is
$\GTwoSThreeEightyFourGain$ ($95\%$ CI
$[\GTwoSThreeEightyFourGainLo,\GTwoSThreeEightyFourGainHi]$) on Gemma-2-2B and
$\GThreeSThreeEightyFourGain$
($[\GThreeSThreeEightyFourGainLo,\GThreeSThreeEightyFourGainHi]$) on Gemma-3-1B,
by bootstrap over latents resampling both designs together, so each gain matches
the difference of its own two rows only to within bootstrap error. Since the
interaction sits in the denominator of $E\rho^2$, the gain column is an
arithmetic consequence of the interaction column, not independent evidence.
Gemma-3-1B's $\GThreeSThreeEightyFourSharedPctAb\%$ is the only shared estimate
here not clipped at the variance boundary; the zeros mean not distinguishable
from zero, not exactly none.}
\label{tab:crossed}
\end{table}

\begin{figure}[t]\centering
\includegraphics[width=\textwidth]{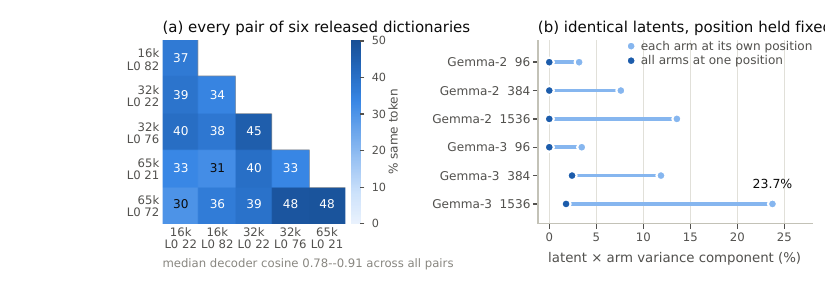}
\caption{(a) Position agreement between every pair of $\ScopeNDicts$ released
Gemma Scope dictionaries at layer $12$, spanning three widths and two sparsity
targets. Latents are matched by mutual-nearest-neighbour decoder cosine at a
fixed threshold and sample size, and the resulting median cosine stays within
$\ScopeCosMin$--$\ScopeCosMax$ for every pair, so cells are comparable and a low
cell is not merely a dissimilar pair. Agreement, averaged over each pair's whole
matched population, runs $\ScopeAgreeMin\%$ to $\ScopeAgreeMax\%$: the
best-agreeing pair of Google's own dictionaries still measures a latent
somewhere else three times in five. Agreement within that population varies with
similarity, from about one in ten at the bottom to
$\RelSparsityTopAgree\%$ among near-duplicates (Appendix~\ref{app:released}). (b) The latent$\times$arm component on identical
latents, with each arm at its own top-activating position (light) and all arms
at one shared position (dark). Connector length is the size of the artifact.
Gemma-3-1B's two larger corpora are the only cells whose shared estimate is not
clipped at the variance boundary; the zeros elsewhere mean not distinguishable
from zero, not exactly none.}
\label{fig:interaction}
\end{figure}

\paragraph{The interaction is the measurement; the coefficient is arithmetic.}
Under per-arm positions over the full sample, the latent$\times$arm component is
$\GTwoSThreeEightyFourUncPctAb\%$ of variance on Gemma-2-2B and
$\GThreeSThreeEightyFourUncPctAb\%$ on Gemma-3-1B. That comparison
is not like-for-like against the shared design, which retains only the latents
firing in every arm at a common position, so we restrict both designs to those
latents: the component is then $\GTwoSThreeEightyFourPerArmPctAb\%$ and
$\GThreeSThreeEightyFourPerArmPctAb\%$ under per-arm positions and
$\GTwoSThreeEightyFourSharedPctAb\%$ and $\GThreeSThreeEightyFourSharedPctAb\%$
under shared positions, on identical latents with
only the measurement location differing (Figure~\ref{fig:interaction}). That is
the result. The generalisability
coefficient rises correspondingly, $\GTwoSThreeEightyFourPerArmErho\to\GTwoSThreeEightyFourSharedErho$ and
$\GThreeSThreeEightyFourPerArmErho\to\GThreeSThreeEightyFourSharedErho$, with
paired gains of $\GTwoSThreeEightyFourGain$ ($95\%$ CI
$[\GTwoSThreeEightyFourGainLo,\GTwoSThreeEightyFourGainHi]$) and
$\GThreeSThreeEightyFourGain$
($[\GThreeSThreeEightyFourGainLo,\GThreeSThreeEightyFourGainHi]$), but since the interaction sits in that coefficient's
denominator, the rise is an arithmetic consequence of the collapse rather than
independent evidence for it, and we report it as such.

Two details belong with those numbers. Gemma-3's
$\GThreeSThreeEightyFourSharedPctAb\%$ is the better estimate
of a small positive component: at smaller evaluation corpora the method-of-moments
estimator returns a negative value and is clipped to zero, so every
smaller-corpus run reports exactly $0.0\%$ and cannot distinguish ``small'' from
``absent''. And the two gains are of comparable size rather than equal: the
intervals are $[\GTwoSThreeEightyFourGainLo,\GTwoSThreeEightyFourGainHi]$ on
Gemma-2-2B against $[\GThreeSThreeEightyFourGainLo,\GThreeSThreeEightyFourGainHi]$
on Gemma-3-1B, which overlap across most of their range while the smaller model's
runs both lower and higher.

\paragraph{The convention does not become less important with more evaluation
data.} Across three corpora spanning a sixteenfold range, position agreement
between arms falls monotonically on both models, single-top-position
agreement $\GTwoSNinetySixTopAgree\%\!\to\!\GTwoSThreeEightyFourTopAgree\%\!
\to\!\GTwoSFifteenThirtySixTopAgree\%$ on Gemma-2-2B and
$\GThreeSNinetySixTopAgree\%\!\to\!\GThreeSThreeEightyFourTopAgree\%\!\to\!
\GThreeSFifteenThirtySixTopAgree\%$ on Gemma-3-1B, with Jaccard falling in step, and the latent$\times$arm component
the convention produces does not shrink with them. The paired gain from
controlling it is $\GTwoSNinetySixGain$, $\GTwoSThreeEightyFourGain$ and
$\GTwoSFifteenThirtySixGain$ on Gemma-2-2B and $\GThreeSNinetySixGain$,
$\GThreeSThreeEightyFourGain$ and $\GThreeSFifteenThirtySixGain$ on
Gemma-3-1B. Extrapolating the two smaller corpora to a realistic evaluation
budget would predict the effect vanishing; the third corpus is what tests that,
and it does not. We claim only that the effect does not disappear at larger
budgets, which is the question a reader deciding whether this matters at scale
is asking. The uncontrolled $E\rho^2$ is non-monotone on both models
($\GTwoSNinetySixUncErho$, $\GTwoSThreeEightyFourUncErho$,
$\GTwoSFifteenThirtySixUncErho$ and $\GThreeSNinetySixUncErho$,
$\GThreeSThreeEightyFourUncErho$, $\GThreeSFifteenThirtySixUncErho$, the middle
point highest in both cases) and we offer no account of that.

\paragraph{The filter cannot manufacture the collapse, and it biases against
us.} The shared design keeps only latents firing in every arm at a common
position, $55\%$ and $35\%$ of those evaluated, which invites the reading
that data was discarded until the result appeared. Two things rule that out.
The filter is applied to both sides: the per-arm and shared components in
Table~\ref{tab:crossed} come from the same retained latents at the same
replication, and a filter applied symmetrically cannot create a difference
between the two designs. And its bias is measurable, running against the claim.
The latents it excludes are the less positionally consistent ones: those
first qualifying at $1536$ sequences have median consistency $\GThreeNewCons$
against $\GThreeOldCons$, and carry $\GThreeNewPctAb\%$ latent$\times$arm
against $\GThreeOldPctAb\%$ (Appendix~\ref{app:composition}). We report on the
population where position matters least, so the component is a lower
bound rather than a flattering subset.

\paragraph{The repair does not depend on how the common position is picked, mostly.} Our shared rule ranks candidates by the minimum activation
across arms, which is what makes it arm-symmetric and also what makes it
select toward tokens where every arm fires hard, and so away from the tokens
the arms most disagree about. That is a way our own rule could be flattering
itself, so we tested it, with the prediction and the decision rule, including
the outcome in which the repair failed, written down before the run.
A second rule was built not to select that way: candidates are the union of
every arm's own top positions, keeping only tokens where all six arms have
nonzero activation (Appendix~\ref{app:union} gives the construction). The
collapse is unchanged on Gemma-2-2B ($\GTwoSThreeEightyFourSharedPctAb\%$ under
either rule) and roughly halved on Gemma-3-1B
($\GThreeSThreeEightyFourSharedPctAb\%$ against $\GThreeUnionPctAb\%$), still
well below that model's per-arm $\GThreeSThreeEightyFourPerArmPctAb\%$. So part
of the controlled estimate on the smaller model does depend on the selectivity
of the rule, and the objection is partly right. We report the range across two
defensible arm-symmetric rules rather than the better half of it.

Part of that growth is the balanced design admitting more, and less
positionally consistent, latents as the corpus grows, and part is not;
Appendix~\ref{app:composition} separates the two on a latent set held fixed
across all three corpora, and reports why Gemma-2's last two fixed-set values
coincide.

The disagreement is itself an instance of the claim rather than a nuisance: if
which token a latent ``is'' about depends on how the dictionary was fitted,
that is the measurement moving, not the latent.

\section{Position accounts for most of what is left}
\label{sec:position}
The within-cell term of Table~\ref{tab:crossed} is position within latent, and
it is $\SixArmPctResid\%$ of variance, larger than latent, arm, and their
interaction combined. This is a result, not a residue: it says the token a
latent is measured at moves the answer more than which dictionary produced the
latent, which is the paper's claim stated as a variance share.

What is not established is which token, and why: regressing this term on
relative position, activation magnitude and within-cell activation rank together
explains $R^2=\PosResidRTwo$ of it, so no obvious covariate substitutes for the
choice. That limits explanation, not the repair: holding position fixed
removes the latent$\times$arm component whether or not we can say what
distinguishes one firing position from another, as randomising any nuisance
factor works without a theory of it. It is also worse for a practitioner than a
monotone trend, which could at least be adjusted for.

Of the five audited papers (Table~\ref{tab:audit}), the ones whose
methodology we could confirm in detail measure each latent at effectively one
position per instance -- a single curated token (Templeton et al.'s case-study
ablations), the top-activating token (the convention we ourselves used before
this decomposition), or an aggregate that is not itself decomposed by position
(Gao et al.'s ablation-sparsity metric). What fraction of a latent's own
measured effect is attributable to which position was chosen is not something
any of them set out to report, and there is no reason they would have: at
$\SixArmPctResid\%$ it is the largest term in the decomposition, and it is
invisible unless the decomposition is run.

\begin{table}[t]\centering
\footnotesize
\begin{tabular}{@{}p{2.6cm}p{4.0cm}p{2.1cm}p{2.6cm}@{}}
\toprule
paper & readout & normalises by magnitude & reports position \\
\midrule
Bricken et al.\ 2023 & per-feature ablation, logit change &, &, \\
\addlinespace[2pt]
Marks et al.\ 2024 & single-latent ablation, KL &, &, \\
\addlinespace[2pt]
Templeton et al.\ 2024 & per-feature zero-clamp &, & one curated token \\
\addlinespace[2pt]
Gao et al.\ 2024 & per-latent ablation, sparsity of effect vector & different axis &, \\
\addlinespace[2pt]
Cho et al.\ 2026 & single-latent ablation, KL/logit score &, &, \\
\bottomrule
\end{tabular}
\caption{The two conventions this paper measures, across the five papers we
could confirm use single-latent zero-ablation with a magnitude readout. Both
columns record what is \emph{reported}, not what was done: ``, '' means the
sections we reviewed do not state it, and we would expect several of these
controls to have been applied without being written up. This describes reporting
practice, ours included: an unreported control cannot be checked or reproduced,
whoever omitted it. Appendix~\ref{app:audit} gives the six-column version and
the four papers excluded as a different readout family.}
\label{tab:audit}
\end{table}

\section{What a defensible measurement needs}
\label{sec:protocol}
Four items, each established above rather than below. A practitioner reporting
an ablation-based causal effect should:
\begin{enumerate}
\item \textbf{Report the position, and how many positions per latent were
measured.} This is the item this paper is about. Measuring at one position per
latent, the common convention, lets the dictionary choose where it is
evaluated, and cannot distinguish a stable causal effect from the token that
happened to be picked (\S\ref{sec:property},~\S\ref{sec:position}).
\item \textbf{Report effect per unit perturbation norm}, not raw. Ablation
perturbs the residual stream by $a_f d_f$, and activation magnitude covaries
with firing frequency, median perturbation norm runs $\PNormRare$ in the
rarest decile to $\PNormFreq$ in the most frequent, so raw effect sizes
compare interventions of unequal size. This alone reverses the sign of a real
comparison: asked whether a dictionary's rarest activation-frequency decile
carries more causal mass than a frequency-matched decile of a tied-random
dictionary, raw KL says worse ($\mathrm{HL}=\RawFlipHL$, $95\%$ CI
$[\RawFlipLo,\RawFlipHi]$, $p=\RawFlipP$) and KL per unit norm says better
($\mathrm{HL}=\NormFlipHL$, $[\NormFlipLo,\NormFlipHi]$, $p=\NormFlipP$), on the
identical latents and the identical control, both intervals excluding one
(Appendix~\ref{app:flip}).
\item \textbf{Drop special-token positions} before any pooled statistic, and
assert the expected explained variance rather than trusting it. The BOS
residual carries an activation the dictionary does not reconstruct, cosine
$0.44$ between input and reconstruction at position $0$ against
$\approx\!0.93$ elsewhere, and its norm dominates any pooled statistic:
including position $0$ moves the explained variance of a released Gemma Scope
dictionary from $0.863$ to $-3.5$.
\item \textbf{Report the range across fitting choices, not a single contrast.}
Any one trained-versus-baseline number, ours included, is one draw from a range
whose width is set by choices the number does not state
(Appendix~\ref{app:endpoints}).
\end{enumerate}
Four further items, on near-duplicate decoder structure, direct-path
unembedding alignment, the sampling scheme, and reporting a rate readout
alongside the magnitude one, affect the size of a measured effect without
deciding its sign. They are stated in
Appendices~\ref{app:dup}--\ref{app:postrend} beside the evidence for each, since
a protocol item whose support a reader cannot see is advice rather than a
requirement.

\section{Limitations and conclusion}
\label{sec:limitations}
\textbf{Two models, one family.} Both base models are Gemma; whether the
convention behaves the same in another architecture is untested.
\textbf{Retained subpopulation.} \S\ref{sec:property} shows the filter cannot
produce the collapse and that it excludes the latents where position matters
most, so the reported component is a lower bound; what it does limit is
external validity, since the controlled numbers describe positionally
consistent latents. The shared design measures
only latents firing in every arm at a common position, $55\%$ and $35\%$ of
those evaluated, and those latents have systematically lower measurement
error than the ones it excludes, so the controlled numbers are claims about
that subpopulation. \textbf{Ablation only.} These
results concern metrics that zero a latent's contribution and read a magnitude;
they do not transfer directly to interchange-based scores such as RAVEL, and we
make no claim about those (\S\ref{sec:relwork}). \textbf{Unexplained.} Why the
within-latent spread across positions is as large as it is, and why five
latents in ninety-three carried an effect that dissolved at four times the
corpus.

\paragraph{Training scale, split into the two claims it affects differently.}
Our arms see $12$M tokens against Gemma Scope's billions, and the natural
objection is that the whole effect is an artifact of undertrained dictionaries.
That objection lands on our two claims very unevenly, and collapsing them into
one limitation would overstate what is at risk.

The \emph{premise}, that two dictionaries select different measurement
positions for the same latent, needs no shared initialisation and no training
of ours, and Figure~\ref{fig:interaction} checks it directly on released
dictionaries: latents matched across a released pair by decoder cosine
\citep{paulo2025sparse}, agreement computed on the matched pairs, across all
$\ScopePairs$ pairs of the grid. The premise is therefore not in question at
production scale; what our own arms add is the controlled comparison, not the
phenomenon.

The \emph{repair}, the crossed decomposition and the $E\rho^2$ gain, cannot be measured this way, and the reason is structural rather than a
shortfall of ambition. A latent-wise variance decomposition requires that
latent $i$ denote the same thing in every arm, which requires a shared
initialisation (\S\ref{sec:setup}), and no released dictionary set provides
one: every public suite varies width, sparsity and seed together. Establishing
the repair therefore requires self-trained arms, and the training budget those
arms can reach on one consumer GPU is the binding constraint. We state this as
a property of the estimand so that a reader does not read the small budget as
carelessness, and so that anyone with more compute knows exactly which
experiment would extend the result: the same six-arm design, one shared seed,
at a production token budget.

\textbf{Corpus-size sensitivity.} Everything here is sensitive to evaluation
corpus size, our own analyses included. Five claims were withdrawn during this
project and two further
mechanisms falsified before reaching a draft, and the pattern is uniform: every
mechanism that survived a $96$-sequence corpus died at $384$, none of them noise
in the usual sense: each was significant, replicated across subsets, and had
a mechanism attached. The results in \S\ref{sec:property} are those that held at
every corpus size on both base models, and any single-corpus effect in this
literature, ours included, should be read as provisional. In each case the check
that caught the effect was imported from an existing standard rather than
produced by re-examining our own numbers.

Five audited papers report a single-latent causal effect with a magnitude
readout; none report the position at which it was taken. Because the convention
selects that position from the dictionary's own activations, two dictionaries
under comparison are measured at different tokens, agreeing on the
top-activating position for $\GTwoSThreeEightyFourTopAgree\%$ and
$\GThreeSThreeEightyFourTopAgree\%$ of latents, less as the corpus grows. Most
of the latent$\times$arm variance
such a comparison attributes to the dictionaries belongs to that instead: on
identical latents it falls from $\GTwoSThreeEightyFourPerArmPctAb\%$ to
$\GTwoSThreeEightyFourSharedPctAb\%$ and from
$\GThreeSThreeEightyFourPerArmPctAb\%$ to
$\GThreeSThreeEightyFourSharedPctAb\%$ once position is fixed. None of this
depends on dictionaries we trained: across $\ScopePairs$ pairs of released Gemma
Scope dictionaries, no pair agrees on the measurement token for more than
$\ScopeAgreeMax\%$ of its matched latents, and even restricting to the latents a
pair encodes almost identically the figure only reaches
$\RelSparsityTopAgree\%$.
The correction is one line of evaluation code, checkable in an afternoon. What
we cannot say is why one firing position differs from another, and that
names the next question: if a latent's effect varies this much across the tokens
where it fires, the quantity to report may be a distribution over those tokens,
an activation-weighted expectation, or a context-conditioned estimand rather
than a scalar, decidable empirically, by whichever is more stable across
dictionaries at equal cost on the crossed design used here.

\section*{Ethics statement}
This paper makes claims about five published papers' reporting, not
about their conduct. Our method, reading methods sections, can establish
what a paper states and cannot establish what its authors did; a control
applied and not written up is invisible to us, and we say so wherever the audit
is cited. Every pipeline variant we compare was written by us from a published
description, so the latitude we document is a property of what a description
can specify, not of anyone's care. Both groups whose results we re-examine
released artifacts that made re-examination possible, which most work of this
kind does not, and that asymmetry should count in their favour rather than
against them. The subject matter is a measurement convention in interpretability
tooling; no models were trained beyond small sparse autoencoders on public text,
no capability was elicited, and no human subjects or personal data are involved.

\section*{Reproducibility statement}
Every quantity in this paper resolves from a results file on disk through one
generator script, and a checker verifies that each macro the manuscript
references is defined and still equals what the source CSVs produce. Two
defects this pipeline had are worth naming, because both changed published
numbers: bootstrap intervals were drawn from a single module-level generator,
so inserting an unrelated quantity upstream silently moved intervals
downstream, and the latent sampler drew from the live-set intersection, so a
rerun whose intersection shifted by one latent drew an almost entirely
different sample. Intervals now seed from their own inputs and the sampler
draws from a fixed identifier space. The second fix took two attempts to land,
and the gap is worth naming because an earlier version of this statement did
not notice it. The sampler was corrected and the corrected evaluation files
were written, but the macro pipeline went on reading the pre-fix files, so a
Gemma-2 quantity at one corpus size was compared against a larger corpus
sharing $7$ of its $240$ latents: a between-sample difference reported as a
corpus effect. Every number so derived has been regenerated, and
\texttt{src/sampler\_check.py} now refuses any comparison between evaluation
files sharing under $80\%$ of their latents. Fixing a defect and repointing the
analysis at the fixed output are separate actions, and only the first had been
done.

Predictions and decision rules were written to a pre-registration log before
the corresponding runs wherever possible, and entries reconstructed after the
fact are marked as such. The final experiment reported here, whether the
smaller model's null was statistical power or a model difference, had its
predictions and decision rule committed to that log before the data were
examined, including a branch distinguishing a genuine effect from estimator
conditioning.

Reproducing the measurement requires no training beyond twelve SAEs: six per
base model, each $12$M tokens on one consumer GPU, about four hours per model.
Base models and Gemma Scope dictionaries are public. The two corpora are
different and both matter: the arms are \emph{trained} on WikiText-103, which
the $12$M-token budget requires, and every \emph{evaluation} in this paper runs
on WikiText-2 (raw), whose $11{,}843$ paragraphs passing our length filter yield
$3{,}432$ distinct $512$-token sequences, about $1.8$M tokens. The corpus ladder
of $96$, $384$ and $1536$ sequences therefore uses at most $45\%$ of what is
available, and no sequence is reused at any rung. Seeds are fixed at $0$
throughout, all six arms share one
initialisation so that latent $i$ denotes the same initial direction, and both
position modes are a single flag on the evaluation script. A seed-$0$ retrain
reproduced the variance components to within $1.8$ percentage points but was
not bitwise identical, and it drew a different latent sample under the
old sampler, so it establishes that the components are stable across an
environment change, not that the pipeline is deterministic. We report it as the
former.

\subsubsection*{Acknowledgements}
This work was supported by a Goodfire Research Grant, and made use of Silico,
Goodfire's research agent.

\bibliographystyle{plainnat}
\bibliography{refs}

\appendix

\section{The sign flip in full}
\label{app:flip}
Protocol item 2 states the reversal; this is the comparison it came from.
Take the single contrast a practitioner would run first: does this dictionary's
rarest activation-frequency decile carry more or less causal mass than a
frequency-matched decile of an untrained, tied-random dictionary? Under raw KL
the rare decile is worse than random ($\mathrm{HL}=\RawFlipHL$, $95\%$ CI
$[\RawFlipLo,\RawFlipHi]$, $n=\RawFlipNT$ against $\RawFlipNR$,
Mann--Whitney $p=\RawFlipP$). Under KL per unit perturbation norm the identical
latents and the identical control are better than random
($\mathrm{HL}=\NormFlipHL$, $95\%$ CI $[\NormFlipLo,\NormFlipHi]$,
$p=\NormFlipP$). Both intervals exclude one. Nothing about the sample, the model
or the dictionary changed between the two readings; only whether the
intervention's own size was accounted for did.

\paragraph{Corollary.} Activation frequency predicts causal mass in neither the
trained nor the control arm once magnitude is accounted for. The apparent
frequency dependence here is magnitude, not frequency, and it would have shipped
as a rarity finding under the more common uncontrolled convention.

\section{Position agreement between released production dictionaries}
\label{app:released}

The premise of this paper does not depend on our own training. It says two
dictionaries select different measurement positions for the same latent, and
that is checkable on dictionaries someone else trained at full scale.

\paragraph{Method.} We take two Gemma Scope dictionaries on the same base model
and layer differing in width or sparsity target, and match latents across them
by decoder cosine, since released dictionaries share no index correspondence.
Matching is mutual nearest neighbour above a stated cosine threshold, and we
report how many pairs survive at each of several thresholds rather than
choosing one: a permissive threshold admits pairs that are not the same
feature, and a strict one selects the most duplicated directions. On the
surviving pairs we compute each dictionary's own top-activating positions under
its own gate and report mean Jaccard overlap and single-top-position agreement,
the same two quantities as \S\ref{sec:property}.

\paragraph{Result.} On pairs matched above cosine $0.70$, median cosine
$\RelSparsityNearMedCos$ and $\RelWidthNearMedCos$, so near-duplicate
directions, the two dictionaries pick the same single top-activating
position for $\RelSparsityNearAgree\%$ and $\RelWidthNearAgree\%$ of latents,
at Jaccard $\RelSparsityNearJaccard$ and $\RelWidthNearJaccard$. Both are far
below agreement, on dictionaries trained by someone else at production scale.

\begin{figure}[t]\centering
\includegraphics[width=\textwidth]{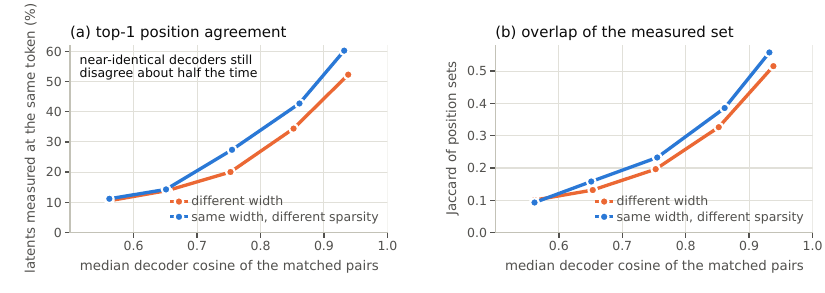}
\caption{Agreement between two released Gemma Scope dictionaries against how
alike the matched latents are, sampled inside five disjoint cosine bands so each
point sits at a stated similarity rather than behind a threshold. (a) share of
latents both measure at the same token; (b) overlap of the six-position sets.
Both rise monotonically with decoder similarity, which is what makes the
near-duplicate figures an upper bound rather than a typical value. Even above
median cosine $0.9$, where the two dictionaries encode a latent almost
identically, they disagree about where to measure a large fraction of the time.}
\label{fig:released}
\end{figure}

\paragraph{Agreement is a distribution, not a point.} Matched pairs are the
latents whose decoder directions already agree most, so they should agree about
position more than a typical pair would. Loosening the threshold does not show
this, since the eligible pool grows but the sample stays dominated by
near-duplicates, so we instead sample \emph{within} disjoint cosine bands
(Figure~\ref{fig:released}). Agreement rises monotonically with decoder
similarity, from $\RelSparsityBandAgree\%$ and $\RelWidthBandAgree\%$ in
$0.50$--$0.60$ to $\RelSparsityTopAgree\%$ and $\RelWidthTopAgree\%$ above
$0.90$.

\paragraph{Which band a reader should have in mind.} The bands are not equally
populated, and reporting the low one alone would describe a tail as though it
were the centre. Across the $\ScopePairs$ pairs, mutual nearest neighbours have
median cosine $\ScopeCosPairMedian$; only $\ScopeShareLow\%$ of them fall in
$0.50$--$0.60$ while $\ScopeShareTop\%$ sit above $0.90$. Weighting each band's
agreement by its share gives $\RelSparsityWeighted\%$ and $\RelWidthWeighted\%$
for a randomly drawn matched pair, consistent with the pair-level figures in
Figure~\ref{fig:interaction} computed a different way. So the modal matched pair
is a near-duplicate, and the honest headline is the one that needs no band at
all: even where two released dictionaries encode a latent almost identically,
they still measure it at different tokens a large fraction of the time, and
agreement never approaches what comparing two published numbers assumes. That is
fatal for a published number without depending on which band we chose.

\paragraph{What this cannot show.} The repair: the variance decomposition needs
a shared initialisation across arms, and no released suite provides one
(\S\ref{sec:limitations}).

\section{A shared-position rule that does not select on agreement}
\label{app:union}
The rule used throughout ranks each latent's candidate positions by the
minimum gated activation across the six arms. That is what makes it
arm-symmetric, no arm serves as the reference, and it is also selective
in a direction that matters: a token where five arms fire hard and one barely
fires ranks below a token where all six fire moderately, so the rule prefers
positions of high inter-arm agreement and avoids the contested ones. Since the
contested positions are what this paper is about, some of the reported
interaction collapse could be that selection rather than the position control
it is attributed to.

\paragraph{The comparator.} For each latent, take the union of every
arm's own top-$n$ activating positions, so each arm's argmax is a candidate
however much the other arms dislike it. From that union keep only the tokens at
which every arm has nonzero activation, without this the design stops
being crossed, and an arm contributing a null effect at a token where it does
not fire would sit at the log floor and dominate the decomposition. Rank the
survivors by the mean activation across arms: the maximum would
privilege whichever arm fires hardest and reintroduce an arm as reference, and
the minimum would rebuild the rule under test. The result is arm-symmetric and
non-selective with respect to inter-arm agreement, while still guaranteeing
that every arm fires where it is measured.

\paragraph{Result.} Retention falls, as it must, contested tokens are where
some arm is more likely silent: from $\GTwoMinGatedRetained$ to
$\GTwoUnionRetained$ latents on Gemma-2-2B and $\GThreeMinGatedRetained$ to
$\GThreeUnionRetained$ on Gemma-3-1B, leaving $\GTwoUnionN$ and $\GThreeUnionN$
in the balanced cube. The latent$\times$arm component under the union rule is
$\GTwoUnionPctAb\%$ on Gemma-2-2B, identical to the minimum-gated
$\GTwoSThreeEightyFourSharedPctAb\%$, and $\GThreeUnionPctAb\%$ on Gemma-3-1B
against $\GThreeSThreeEightyFourSharedPctAb\%$. Paired gains are
$\GTwoUnionGain$ ($95\%$ CI $[\GTwoUnionGainLo, \GTwoUnionGainHi]$) and
$\GThreeUnionGain$ ($[\GThreeUnionGainLo, \GThreeUnionGainHi]$).

\paragraph{Reading it.} On Gemma-2-2B the two rules are indistinguishable, so
candidate selection is not what removes the interaction there. On Gemma-3-1B
roughly half the collapse survives the change of rule; the remainder was
selection. Gemma-3-1B's gain is not distinguishable from zero under the union
rule at $\GThreeUnionN$ latents, which we cannot separate from the loss of
power that the stricter retention causes. This was pre-registered with its
decision rule before the union evaluation ran, including the outcome in which
the repair failed entirely; the outcome that occurred was a partial one on one
model and a clean survival on the other, and both are reported here rather than
the better of them. What it does not establish is that \emph{every}
arm-symmetric rule agrees, two rules agreeing is two data points, and
neither is derived from a definition of what the estimand should be.

\section{How much of the corpus growth is composition}
\label{app:composition}
The paired gain grows with evaluation corpus on both models
(\S\ref{sec:property}). Two things could produce that: the effect strengthening
with data, or the balanced design admitting different latents at different
corpus sizes. The second is real and measurable. On Gemma-3-1B the
$\GThreeNewN$ latents that first qualify at $1536$ sequences have median
positional consistency $\GThreeNewCons$ against $\GThreeOldCons$ for those
already balanced at $96$, and carry $\GThreeNewPctAb\%$ latent$\times$arm
against $\GThreeOldPctAb\%$; on Gemma-2-2B the corresponding figures are
$\GTwoNewCons$ against $\GTwoOldCons$ and $\GTwoNewPctAb\%$ against
$\GTwoOldPctAb\%$. The newcomers are the sparser, less positionally consistent
latents, exactly where position selection has most room to disagree.

Restricting all three corpora to the latents balanced in every one of them
holds the sample fixed, so any remaining trend is about data. The gain then
reads $\GThreeSNinetySixFixedGain$, $\GThreeSThreeEightyFourFixedGain$ and
$\GThreeSFifteenThirtySixFixedGain$ on $\GThreeFixedN$ Gemma-3-1B latents, and
$\GTwoSNinetySixFixedGain$, $\GTwoSThreeEightyFourFixedGain$ and
$\GTwoSFifteenThirtySixFixedGain$ on $\GTwoFixedN$ Gemma-2-2B latents. The
growth is therefore part composition and part not, in proportions that differ
by model, which is why \S\ref{sec:property} claims only that the effect does
not vanish rather than that it strengthens.

\paragraph{The Gemma-2 tie at 384 and 1536.}
The fixed-latent gain reads $+0.110$ at both $384$ and
$1536$ sequences on the same $\GTwoFixedN$ latents, which invites reading as a
ceiling. It is not one. The components underneath the two rows differ
substantially, the per-arm interaction rises $0.0432 \to 0.0539$ ($+25\%$),
the shared design's between-latent variance falls $0.3383 \to 0.2757$
($-19\%$), and residual variance falls in both designs, while the two
coefficients they produce happen to land $0.001$ apart
($0.8879-0.7760 = +0.1119$ against $0.8738-0.7630 = +0.1108$). A tie in a
difference between two moving quantities is arithmetic, not a plateau, and the
bootstrap interval at either corpus is more than a hundred times the gap. We
report the fixed-set gain as rising from the smallest corpus and unchanged
thereafter, and claim nothing about a ceiling.

\section{AI use statement}
\label{app:aiuse}
This work was carried out with assistance from AI coding agents:
 Claude Code, and Silico,
Goodfire's research agent. The agents wrote and ran analysis code,
executed the experiments described here on the author's hardware, drafted prose,
and audited the literature by reading source PDFs. All experimental designs,
pre-registered predictions and decision rules were fixed before the
corresponding runs and recorded in a pre-registration log maintained alongside
the analysis. The author is responsible for all claims, verified the numerical
results against the generated-macro pipeline, and confirmed each cited paper's
methodology against the primary source.

\section{The per-arm decomposition that produced the retracted claim}
\label{app:olddecomp}
This paper originally reported the decomposition below and concluded from it
that causal importance is not a property of the latent. Every arm was measured
at its own top-activating positions, so position was nested within arm; the
latent$\times$arm term therefore mixes the latent behaving differently under a
different fit with the arms having chosen different tokens, and the two are not
separable after the fact. It is reproduced here for completeness, not as a
result.

\begin{center}
\begin{tabular}{lcc}
\toprule
component & 2 arms (extrapolated) & 6 arms (direct) \\
\midrule
latent                        & --                   & $\SixArmPctLatent\%$ \\
arm                           & $\TwoArmPctArm\%$    & $\SixArmPctArm\%$ \\
latent $\times$ arm           & $\TwoArmPctInter\%$  & $\SixArmPctInter\%$ \\
within-cell (position)        & --                   & $\SixArmPctResid\%$ \\
$E\rho^2$ (one-arm design)    & $\TwoArmErhoTwo$     & $\SixArmErhoTwo$ \\
\bottomrule
\end{tabular}
\end{center}

\noindent Gemma-2-2B, $96$-sequence evaluation corpus, $\SixArmI$ latents
$\times$ $\SixArmJ$ arms $\times$ $\SixArmN$ positions per cell. The two-arm
column is an earlier trained-versus-soft-frozen design carrying one degree of
freedom on the arm factor. Table~\ref{tab:crossed} supersedes both columns.

\section{Near-duplicate structure}
\label{app:dup}
\textbf{Mechanism.} Feature splitting produces decoder rows at high mutual
cosine; ablating one member of such a pair leaves the concept partly represented.
\textbf{Evidence.} Per-latent maximum cosine predicts causal mass in the released
Gemma Scope dictionary ($\rho=+0.220$, $p=0.0006$) and in neither of our trained
arms ($+0.078$, $+0.060$), which have visibly less splitting (medians $0.207$ and
$0.122$ against $0.272$). \textbf{Control.} One-sided adjustment where the
relationship is asymmetric. Symmetric residualisation is wrong here: fitting the
control arm produces a slope of $+2.506$ against a predictor with $p=0.82$, i.e.\
residualising against noise. Matching is infeasible against a random dictionary,
which structurally cannot contain near-duplicates.

\paragraph{This artifact is conditional.} Near-duplicate structure confounds
causal-effect measurement in dictionaries that have it. That is more
useful than a universal claim, because it comes with its own diagnostic.

\paragraph{Protocol item.} Report the per-latent nearest-neighbour cosine
distribution, and adjust or match on it \emph{in dictionaries that exhibit
near-duplicate structure}. Whether yours does is one line of code.

\section{Direct-path alignment: a sensitivity, not a confound}
\label{app:align}
\label{sec:align}
\textbf{Mechanism.} A trained decoder direction is aligned with directions the
model writes to the residual stream, and those feed the unembedding. Next-token
KL therefore rewards latents that short-circuit the network. We measure alignment
as $\sqrt{d^\top C d}$ with $C$ the embedding covariance: the spread of the
direct logit contribution, since a uniform logit shift leaves the softmax
unchanged.

\textbf{Evidence.} Alignment predicts causal mass at $\rho=+0.375$ in Gemma Scope
and $+0.264$ in a dictionary we trained ourselves with a different architecture
and roughly $40\times$ less data, with the control arm null in both cases
($-0.063$, $+0.004$). The single most causally powerful latent we found writes the
letter \emph{k} to the logits and carries $12\times$ the causal mass of the
runner-up.

\textbf{Why this is a sensitivity and not a confound.} The trivially aligned
latents are approximately the top $20$ of $16{,}384$, the $99.88$th percentile. No
sample a paper would realistically draw reaches them: the top twenty by alignment
within a $240$-latent sample sit near the $92$nd dictionary percentile and are
$90\%$ word-like, under both stratified and uniform sampling. So at realistic
sample sizes high-alignment latents are semantic, and adjusting alignment away
removes signal rather than triviality. The correct statement is that
ablation-based measurements are sensitive to unembedding alignment at
$\rho\approx0.3$, and that within realistic samples this alignment reflects
semantic content.

\textbf{Readouts disagree, and the disagreement is ordered.} Ablation,
interchange and clamping all perturb along the same direction $d_f$ and differ
only in the scalar, so ``use interchange instead'' is not a remedy. The axis
that matters is magnitude versus rate: a magnitude readout scales with
alignment, whereas a rate is a threshold crossing and saturates once the
intervention flips the answer. Computing all of them from identical forward
passes across the layer-12 dictionaries ($n=\RateN$), alignment predicts

\begin{center}
\begin{tabular}{lc}
\toprule
readout & $\rho$ with alignment \\
\midrule
KL per unit perturbation norm (unbounded magnitude) & $\RateRhoKL$ \\
KL, raw (unbounded magnitude)                       & $\RateRhoKLRaw$ \\
$|\Delta p|$ of the original top token (bounded magnitude) & $\RateRhoAdtop$ \\
argmax changed (rate)                               & $\RateRhoFlip$ \\
\bottomrule
\end{tabular}
\end{center}

\noindent all significant, and ordered as the saturation account predicts:
unbounded magnitude, then bounded magnitude, then rate. The end-to-end factor
between the magnitude readout in common use and the rate readout is
$\RateRatio$.

\textbf{Why we report this as a disagreement and not yet as a
recommendation.} Measured flip rates run $\RateFlipMin$--$\RateFlipMax$, so the
rate readout is a rare event in this design and has genuinely less statistical
power than KL. Part of its smaller correlation is therefore attenuation rather
than saturation, and these data cannot separate the two: a design with flip
rates near $0.5$ could. The ordering above is evidence for saturation, since
$|\Delta p|$ is not a rare event and still lands between, but it is not
decisive. We therefore recommend reporting both readouts (\S\ref{sec:protocol})
rather than preferring one, and record the choice between them as open
(\S\ref{sec:limitations}).

\textbf{A caution about the signed readout.} The
probability change of the original top token is \emph{signed}, and
$\RateDtopNegPct\%$ of our measurements are negative: ablating a latent
frequently \emph{raises} the top token's probability. A rank correlation
computed against the signed quantity measures direction rather than size, and
its per-latent median is near zero; read that way it gives
$\rho=\RateRhoDtopSigned$ and appears to refute the ordering above. It does not: it is not a magnitude. The magnitude is $|\Delta p|$.

\textbf{An unexplained depth dependence.} The measure is layer-independent by
construction, so it should predict best where fewest blocks intervene, deepest. It does the opposite: pooling by Fisher $z$, alignment predicts causal
mass at $\rho=\RhoShallow$ shallow against $\RhoDeep$ deep, a contrast of
$z=\RhoContrastZ$ ($p=\RhoContrastP$). We report this as unexplained; it is a
violation of the measure's own construction rather than an observation about
the model. Appendix~\ref{app:depth} gives the per-layer estimates, the trend
test, and the falsification of the compressed-variance explanation.

\paragraph{Protocol items.} Report unembedding alignment alongside the effect,
giving both the unadjusted and alignment-adjusted estimates, since this is a
sensitivity rather than a confound. State the sampling scheme: the size of the
sensitivity depends on it, and no sample of a few hundred latents reaches the
extreme tail of the alignment distribution. And report a rate readout alongside
the magnitude one, because they disagree by a factor of $\RateRatio$ on
identical forward passes.

\section{Position in sequence}
\label{app:postrend}
\textbf{Mechanism.} Effects are measured at each latent's top-activating
positions, and later positions carry more context and a sharper next-token
distribution. \textbf{Evidence.} Pooled $\rho=-0.087$ ($p<10^{-4}$) across
$2{,}700$ latents, reaching $-0.207$ in one dictionary and null in five. The sign
is negative: earlier positions carry slightly more causal mass, opposite to
the mechanism we proposed. On the endpoint sample the effect is not significant
($\rho=-0.084$, $p=0.196$). This correlational null does not contradict
\S\ref{sec:position}'s variance share: a latent's effect can vary substantially
across its own top positions ($\SixArmPctResid\%$ of variance) without that
variation trending consistently early-to-late across latents ($\rho\approx0$).
\textbf{Control.} Record and report position, and how many positions per latent
were sampled; a single position per latent cannot distinguish a stable causal
effect from a coincidence of which token was chosen.

\section{Endpoints}
\label{app:endpoints}
Table~\ref{tab:endpoints} gives the trained-versus-control comparison for three
controls. The second and third rows share a dictionary, a feature list and an
evaluation, so their ratio is a within-experiment decomposition.

\begin{table}[t]\centering
\begin{tabular}{lccc}
\toprule
comparison & HL & 95\% CI & $n$ \\
\midrule
Gemma Scope vs untrained tied & $\GSvsRandHL$ & $[\GSvsRandLo, \GSvsRandHi]$ & $\GSvsRandN$ \\
ours vs untrained tied        & $\MineVsUntrHL$ & $[\MineVsUntrLo, \MineVsUntrHi]$ & $\MineN$ \\
ours vs soft-frozen           & $\MineVsFrozenHL$ & $[\MineVsFrozenLo, \MineVsFrozenHi]$ & $\MineN$ \\
\bottomrule
\end{tabular}
\caption{Causal effect per unit perturbation norm, Hodges--Lehmann ratios on
uniform samples of live latents. The soft-frozen control has a trained encoder
and a decoder constrained to cosine $0.8$ of its random initialisation.}
\label{tab:endpoints}
\end{table}

A trained dictionary beats an untrained one by $\MineVsUntrHL\times$ and a
soft-frozen baseline by $\MineVsFrozenHL\times$ ($p=\MineVsFrozenP$). Most of the
measured advantage is therefore attributable to the encoder; decoder freedom
contributes the smaller factor. This holds despite the two arms differing sharply
in decoder geometry, median cosine to initialisation $0.231$ for the free
decoder against $0.800$ for the constrained one, and despite a reconstruction
gap that grows monotonically over training ($+0.0064$ explained variance per
million tokens, $p<10^{-4}$).

\paragraph{The matched estimand, and the spread around it.} Among the six arms,
four leave the decoder free (\texttt{free}, \texttt{lr1e4}, \texttt{order1},
\texttt{k41}) and two soft-freeze it (\texttt{tau080}, \texttt{tau090}). Only
\texttt{free} shares learning rate, sparsity and data order with the frozen
arms, so only \texttt{free} vs \texttt{tau080} and \texttt{free} vs
\texttt{tau090} isolate decoder freedom. Those are the estimand:
\[
\MatchedTauEightyHL\times\ [\MatchedTauEightyLo, \MatchedTauEightyHi]
\quad\text{and}\quad
\MatchedTauNinetyHL\times\ [\MatchedTauNinetyLo, \MatchedTauNinetyHi],
\]
paired Hodges--Lehmann on the same latents, against the two freeze strengths.
The first interval includes $1$; the second does not. Even the matched
comparison therefore does not give a single stable answer: it gives an answer
that depends on how hard the baseline is frozen.

The remaining six contrasts vary a second factor and are a sensitivity, not the
estimand. Across all $\ContrastAllN$, paired ratios span $\ContrastAllLo\times$
to $\ContrastAllHi\times$ with $\ContrastAllSigN$ intervals excluding $1$;
dropping \texttt{k41} leaves $\ContrastLo$--$\ContrastHi\times$, median
$\ContrastMedian\times$, $\ContrastSigN$ of $\ContrastNoKFourOneN$ excluding
$1$. The \texttt{k41} contrasts are the largest in the set for a reason worth
stating rather than hiding: at $k{=}41$ the same reconstruction is carried by
half as many active latents, so ablating one removes more of it. That is a
sparsity effect masquerading as a training effect, and it is exactly the kind of
uncontrolled comparison this paper is about. One contrast (\texttt{lr1e4} vs
\texttt{tau080}) sits at $\ContrastLo\times$: no advantage at all.

We report the matched pair as the result and the spread as its sensitivity,
rather than the most favourable member of either.

\paragraph{Why the soft-frozen baseline is the right comparison even though
nobody trains one.} A reviewer may object that soft-freezing at cosine $0.8$ to
initialisation is not a choice any practitioner makes, so beating it is not
informative. The construction is deliberate: it is close to the
smallest perturbation that still yields a genuinely different dictionary,
since the decoder is pinned near its random start while the encoder trains
freely. Dictionaries that differ for the reasons real dictionaries differ, seed, corpus, width, sparsity, architecture, differ by more than this. So the
instability measured against this baseline is a \emph{lower bound} on the
instability a practitioner faces, and the endpoint range above is correspondingly
conservative.

\section{The tail is not a population phenomenon}
\label{app:tail}
Across $\NDicts$ Gemma Scope dictionaries spanning two depths, three widths and
sparsity from $L_0$ $22$ to $445$, the ratio of maximum to median causal mass
within a fixed sample of $300$ latents runs from $\TailMaxMedMin\times$ to
$\TailMaxMedMax\times$, and the top $5\%$ of latents carry between
$\TailMassMin\%$ and $\TailMassMax\%$ of total causal mass. Every dictionary shows
it.

But within any one dictionary the tail is a handful of latents. Removing the
single largest takes the variance ratio from $\SDRatioAll\times$ to
$\SDRatioDropOne\times$; the median ratio is unmoved. A bootstrap resamples those
latents in and out while still treating them as population draws, so the jackknife
curve is the honest instrument. At $n\approx200$ the control arm's $99$th
percentile has a bootstrap interval spanning $11.3\times$, and the derived
``fraction above the control's $p99$'' runs $[4.2\%, 39.6\%]$: published aggregate
comparisons on a few hundred latents cannot distinguish anything.

\paragraph{Protocol item.} Report the jackknife curve, not the mean or the
variance ratio. Removing one latent of $\GSvsRandN$ moves our variance ratio
from $\SDRatioAll\times$ to $\SDRatioDropOne\times$.

\section{Effect decays with remaining depth, but not proportionally}
$\mathrm{KL}_{t+3}/\mathrm{KL}_t$ declines across four depths by a factor of
$\DepthSpan$, which is large and not in doubt. The attribution to propagation
limits is much weaker: of three pre-registered predictions, the power law and
both thresholds were falsified and only bare monotone ordering survived, worth
$p\approx\DepthOrderP$ on four points. The decline is solid, the mechanism is
not established. Because the gradient is a propagation effect it is not evidence
that deeper layers are more direct-path dominated and must not be read as
support for Appendix~\ref{app:align}. Appendix~\ref{app:decay} gives the per-depth
ratios and the pre-registered predictions with their misses.

\section{Reconstruction improves without a matching causal gain}
\label{app:curve}
Evaluating the trained and soft-frozen arms at four training budgets gives HL
ratios of $\CurveHLThreeM$, $\CurveHLSixM$, $\CurveHLNineM$ and $\CurveHLOneTwoM$ at $3$, $6$,
$9$ and $12$ million tokens (slope $\CurveSlope$ per million, $p=\CurveSlopeP$),
while the explained-variance gap over the same range grows from $0.000$ to
$0.077$ at $p<10^{-4}$. The reconstruction gap grows significantly and the causal
gap does not. With four budgets, $120$ latents each, and non-monotone point
estimates, this is consistent with a dissociation but does not establish one.

\section{How Figure 1's latent was selected}
\label{app:figsel}
Of the $131$ latents live in both designs on Gemma-2-2B at the $384$-sequence
corpus, a larger set than the $\GTwoSThreeEightyFourPairedN$ of
Table~\ref{tab:crossed}, which additionally requires six positions per arm so
that the two cubes are balanced, $114$ have differing argmaxes across the two arms shown and $111$ have
an assigned shared position distinct from both. Drawing a single-sequence trace
adds three further requirements: both arms must fire on one common sequence,
the shared position must lie at least four tokens from each argmax so that the
figure does not appear to defer to one arm, and all three tokens must fit in a
readable window. Two latents satisfy all of them; Figure~\ref{fig:mechanism}
shows the first.

\paragraph{Is the latent shown unusually sparse?} It fires at $5$ and $4$ of
$511$ positions on its sequence. We are not able to give a percentile for that,
and say so rather than quoting one: firing density is a property of the
latent--sequence pair rather than of the latent, so a reference
distribution measured on any single sequence is not the right comparison. One
latent we traced fires at $55$ and $58$ positions on its own sequence and at
none at all on another. Since Figure~\ref{fig:mechanism}'s latent was selected
partly because it fires on the sequence shown, comparing it against densities
measured on that same sequence would be a selection effect. The honest
statement is that sparsity varies by orders of magnitude across
latent--sequence pairs and the figure shows one pair.

One caution about the counts above. The evaluation files record each arm's
top-$n$ positions globally, not per sequence, so a within-sequence
argmax read off those files is only the highest recorded position on
that sequence. The figure's argmaxes are therefore taken from the full traces,
and for the latent shown they agree with the recorded values; for a latent we
examined earlier they did not, which is why the counts here are stated for the
global comparison rather than the per-sequence one.

\section{The full literature audit}
\label{app:audit}
Table~\ref{tab:audit} in the body gives the two columns corresponding to the two
conventions the paper measures. The full audit also recorded near-duplicate
structure and unembedding alignment: no paper reports either in a way tied to
its causal-effect measurement, though Bricken et al.\ and Templeton et al.\
discuss feature splitting broadly across dictionary sizes, and Cho et al.\
report a per-layer $\Delta$-logit structure whose relation to unembedding
alignment as defined here we could not establish. Four further papers were
considered and excluded as a different readout family rather than as negative
findings: Makelov et al.\ 2024 (edits 2/4/6 features simultaneously), SAEBench
TPP/SCR (set ablation of several latents per concept), Korznikov, Galichin et
al.\ 2026 (RAVEL interchange match-rate), and Rajamanoharan et al.\ 2024
(whole-dictionary substitution). Exclusions are recorded so that the denominator
of ``none of five'' is visible.

\section{Depth dependence of the alignment sensitivity}
\label{app:depth}
Alignment predicts causal mass with $\rho=\RhoLFive$ at layer 5 and
$\RhoLOneTwo$ at layer 12, against $\RhoLOneNine$ at layer 19 and
$\RhoLTwoFour$ at layer 24 ($n=\DepthNLFive$ per depth).

At $n=300$ a Spearman correlation of $0.3$ carries a $95\%$ interval of about
$\pm\RhoCIHalf$, so the two shallow estimates are indistinguishable from each
other and so, marginally, are the two deep ones. We therefore test the trend
rather than describe the sequence: pooling by Fisher $z$ gives
$\rho=\RhoShallow$ for the shallow pair and $\rho=\RhoDeep$ for the deep pair, a
contrast of $z=\RhoContrastZ$ ($p=\RhoContrastP$), and a weighted regression of
$z$ on layer index has slope $\RhoSlopePerLayer$ per layer where the measure's
construction predicts a positive slope. The pattern is a decrease with depth,
not a mid-network dip with recovery.

The compressed-variance explanation is falsified: alignment spread is equal
across depths (CV $0.1175$ against $0.1091$) and causal-mass spread is
greater at layer 19.

\section{Effect decay with remaining depth: the pre-registered predictions}
\label{app:decay}
Measuring \eqref{eq:kl} at $t$ and $t{+}3$ across four depths at matched width
and approximately matched sparsity gives $\mathrm{KL}_{t+3}/\mathrm{KL}_t$ of
$\DepthRatioLFive$ (layer 5, $20$ blocks remaining), $\DepthRatioLOneTwo$
(layer 12, $13$), $\DepthRatioLOneNine$ (layer 19, $6$) and
$\DepthRatioLTwoFour$ (layer 24, $1$), with $n=\DepthNLFive$ latents per depth.

We pre-registered three predictions of decreasing strength. A power law in
remaining depth, fitted from layers 12 and 19 and tested out of sample with a
factor-of-$1.5$ tolerance, predicts $0.760$ at layer 5 and $0.003$ at layer 24
against observed $\DepthRatioLFive$ and $\DepthRatioLTwoFour$, misses of
$0.51\times$ and $8.4\times$, in opposite directions, so the curve is flatter
than proportional at both ends. Two threshold predictions, $>\!0.5$ at layer 5
and $<\!0.02$ at layer 24, also both failed. Only bare monotone ordering
survived, worth $p\approx\DepthOrderP$ on four points under a
random-permutation null.

\end{document}